\documentclass[conference]{IEEEtran}
\IEEEoverridecommandlockouts
\usepackage{cite}
\usepackage{amsmath,amssymb,amsfonts}
\usepackage{algorithmic}
\usepackage{graphicx}
\usepackage{textcomp}
\usepackage{xcolor}
\def\BibTeX{{\rm B\kern-.05em{\sc i\kern-.025em b}\kern-.08em
    T\kern-.1667em\lower.7ex\hbox{E}\kern-.125emX}}
\usepackage{makecell}
\usepackage{float}
\usepackage{array}
\usepackage{multirow}
\usepackage{booktabs}
\usepackage{pifont}
\usepackage{bm}
\usepackage{url}
\usepackage[hidelinks]{hyperref}
\usepackage{tabularx}
\usepackage{fancyhdr}

\fancypagestyle{firstpage}{
  \fancyhf{}

  \fancyfoot[C]{%
    \parbox{\textwidth}{%
      \centering
      \footnotesize
      \copyright~2026 IEEE. Personal use of this material is permitted.
      Permission from IEEE must be obtained for all other uses, in any
      current or future media, including reprinting/republishing this
      material for advertising or promotional purposes, creating new
      collective works, for resale or redistribution to servers or lists,
      or reuse of any copyrighted component of this work in other works.
    }%
  }
}

\definecolor{CheckBlue}{RGB}{26,133,255}   
\definecolor{CrossRed}{RGB}{212,17,89}     

\newcommand{\cmark}{{\color{CheckBlue}\ding{51}}}
\newcommand{\xmark}{{\color{CrossRed}\ding{55}}}

\newcommand{\mae}{\text{MAE}}
\newcommand{\mape}{\text{MAPE}}

\begin{document}

\title{Benchmarking Time Series Foundation Models for Load Forecasting Under Covariate Uncertainty}

\author{\IEEEauthorblockN{Tomas Kaljevic, Ivan Arzola, Yu Zhang}
\IEEEauthorblockA{Department of Electrical and Computer Engineering \\
University of California, Santa Cruz \\
\{tkaljevi, iarzola, yzhan419\}@ucsc.edu}}

\maketitle
\thispagestyle{firstpage}

\begin{abstract}
Accurate short-term load forecasting (STLF) is essential for the reliable and efficient operation of modern power systems. While time series foundation models (TSFMs) have recently demonstrated remarkable performance across a wide range of forecasting tasks, their effectiveness for STLF under realistic operational conditions remains largely unexplored. In this paper, we present a comprehensive benchmark of four trained-from-scratch (TFS) models and four TSFMs across three real-world load forecasting datasets under operational scenarios that differ in the availability and quality of future covariate information. Our results show that Chronos-2 consistently achieves state-of-the-art performance in both zero-shot and fine-tuned settings when future covariates are available or accurately forecast. However, its performance degrades as covariate forecasts become increasingly noisy, whereas TimesNet exhibits greater robustness under severe covariate uncertainty. These findings demonstrate the effectiveness of covariate-informed TSFMs for STLF while highlighting the critical role of robust covariate modeling in real-world forecasting applications.
\end{abstract}

\section{Introduction}
Grid operators and utility companies rely on accurate load forecasting methods to make informed decisions about network planning and maintenance, and improve resource utilization, energy storage, and grid efficiency~\cite{ullahShortTermLoadForecasting2024}. Moreover, modern power systems have grown in complexity, accommodating an increased penetration of renewable energy sources, a larger share of electric vehicles, and greater power demands associated with the accelerated buildout of data centers and AI infrastructure~\cite{pinheiroShorttermElectricityLoad2023,USDataCenter}. As such, accurate STLF models are needed to facilitate overall system functionality, improve financial planning, and mitigate blackouts and outages.

\begin{table*}[t]
\centering
\caption{Comparison of TSFM capabilities. $^\dagger$TimesFM~2.5 supports exogenous variables through an optional XReg module, which is external to the foundation model and therefore not considered native covariate modeling.}
\label{tab:tsfm_comparison}
\resizebox{\linewidth}{!}{%
\begin{tabular}{lccccccc}
\toprule
\textbf{Model}
& \makecell{\textbf{Univariate}\\\textbf{Forecasting}}
& \makecell{\textbf{Covariate-Informed}\\\textbf{Forecasting}}
& \makecell{\textbf{Past Covariate}\\\textbf{Support}}
& \makecell{\textbf{Future Covariate}\\\textbf{Support}}
& \makecell{\textbf{Quantile Levels}}
& \makecell{\textbf{Max. Context}\\\textbf{Length}}
& \makecell{\textbf{Parameter}\\\textbf{Count}} \\
\midrule
TimesFM 2.5$^\dagger$ & \cmark & \xmark & \xmark & \xmark & $\{0.1, 0.2, \ldots, 0.9\}$ & 16,384 & 200M \\
Chronos-2             & \cmark & \cmark & \cmark & \cmark & $\{0.01, 0.05, 0.1, \ldots, 0.9, 0.95, 0.99\}$ & 8,192 & 120M \\
Moirai 2.0            & \cmark & \xmark & \xmark & \xmark & $\{0.1, 0.2, \ldots, 0.9\}$ & 8,192 & 11.4M \\
TiRex                 & \cmark & \xmark & \xmark & \xmark & $\{0.1, 0.2, \ldots, 0.9\}$ & 2,048 & 35M \\
\bottomrule
\end{tabular}}
\end{table*}

Deep time series forecasting models have achieved strong performance across a wide range of applications, including energy forecasting~\cite{wangDeepTimeSeries2026}. We consider four representative models: DLinear, PatchTST, Informer, and TimesNet. DLinear is a lightweight linear model that decomposes each variate into trend and seasonal components before learning a linear mapping to future values~\cite{zengAreTransformersEffective2022}. As a simple yet competitive alternative to Transformer-based architectures, it serves as a strong baseline. PatchTST is an encoder-only Transformer that divides each input sequence into overlapping patches and models dependencies among them to generate forecasts~\cite{nieTimeSeriesWorth2023}. Informer is an encoder-decoder Transformer that employs sparse self-attention to efficiently capture long-range temporal dependencies~\cite{zhouInformerEfficientTransformer2021}. In contrast, TimesNet is a CNN-based architecture that exploits the multi-periodic nature of time series by transforming one-dimensional sequences into two-dimensional representations via the Fourier transform, followed by convolutional feature extraction~\cite{wuTimesNetTemporal2DVariation2023}. Throughout this paper, models trained exclusively on the target dataset, without pre-training, are referred to as trained-from-scratch (TFS) models.

Recently, time series foundation models (TSFMs) have emerged as a powerful alternative to training models from scratch for specific tasks. TSFMs are pre-trained on a large and diverse corpus of both synthetic and real-world time series data from various domains, and can be used for zero-shot forecasting or adapted for any given downstream task through fine-tuning. We consider four publicly available TSFMs that achieve state-of-the-art performance across real-world benchmarks: TimesFM 2.5~\cite{dasDecoderonlyFoundationModel2024}, Chronos-2~\cite{ansariChronos2UnivariateUniversal2025}, Moirai 2.0~\cite{liuMoirai20When2026}, and TiRex~\cite{auerTiRexZeroShotForecasting2025}. Except for TiRex, which is based on the xLSTM architecture~\cite{beckXLSTMExtendedLong2024}, each of these models are based on the Transformer architecture, and each incorporate techniques found in many deep time series models, including patching, embedding, and tokenization. Each model supports both point forecasting and quantile forecasting. Among these models, we distinguish between two forecasting paradigms relevant to our task: univariate forecasting, where only historical load demand is used for forecasting, and covariate-informed forecasting, where exogenous variables like weather conditions and calendar effects are used to forecast future load demand. The capabilities of each model are summarized in Table~\ref{tab:tsfm_comparison}.

Previous studies have identified TSFMs as a promising approach for accurate STLF. Saravanan et al. and Meyer et al. benchmarked several TSFMs, including earlier versions of some models evaluated in this work, against classical and deep learning-based TFS models for commercial and residential building load forecasting~\cite{saravananAnalyzingPerformanceTime2024,meyerBenchmarkingTimeSeries2025}. Both studies found that TSFMs achieve performance comparable to, and in some cases exceeding, TFS models. However, neither investigated TSFM fine-tuning or the use of future covariates. Bose et al. showed that fine-tuning TimesFM 1.0 for commercial building load forecasting outperformed all evaluated TFS models~\cite{bose2024from}. Kreusel et al. investigated covariate integration by incorporating outdoor temperature and calendar information for office building load and cooling demand forecasting~\cite{kreuselCovariatesAreKey2026}. However, their study did not compare TSFMs against a broad range of TFS models under the same forecasting tasks.

\textit{Contributions}. We present a comprehensive benchmark comparing four TFS models (DLinear, PatchTST, Informer, and TimesNet) and four time series foundation models (TSFMs) (TimesFM 2.5, Chronos-2, Moirai 2.0, and TiRex) for day-ahead load forecasting. 
We evaluate all models across three real-world datasets, including a newly constructed CAISO dataset that integrates load demand and weather covariates.
Our evaluation considers three practical operational scenarios reflecting different levels of future covariate availability: (i) unavailable, (ii) perfectly known, and (iii) available only as noisy forecasts. Across these scenarios, Chronos-2 consistently achieves state-of-the-art performance in both the zero-shot and fine-tuned settings, while TimesNet exhibits greater robustness under severe covariate uncertainty.

\section{Experiment Design}
\label{sec:exp_design}

The multi-step time series forecasting problem can be formulated either as learning a deterministic mapping
\begin{align*}
    \hat{y}_{t+1:t+H}
    = f\!\left(
    y_{t-L+1:t},
    \mathbf{x}_{t-L+1:t},
    \hat{\mathbf{x}}_{t+1:t+H};
    \bm{\theta}
    \right),
\end{align*}
for point forecasting, or as learning the conditional distribution
\begin{align*}
    \mathcal{P}\!\left(
    y_{t+1:t+H}
    \mid
    y_{t-L+1:t},
    \mathbf{x}_{t-L+1:t},
    \hat{\mathbf{x}}_{t+1:t+H};
    \bm{\theta}
    \right),
\end{align*}
for probabilistic (quantile) forecasting, where $\bm{\theta}$ denotes the learnable model parameters. In our experiments, $y_t \in \mathbb{R}$ represents the load demand at time $t$, $\mathbf{x}_t \in \mathbb{R}^p$ denotes the vector of external covariates, and $L$ and $H$ are the context length and prediction horizon, respectively. Depending on the forecasting model, the inputs may include the historical target sequence $y_{t-L+1:t}$, the historical covariate sequence $\mathbf{x}_{t-L+1:t}$, and forecasts of the future covariates $\hat{\mathbf{x}}_{t+1:t+H}$. However, support for future covariates is model-dependent and not universally available.

\subsection{Datasets}

We evaluate each model on three publicly available datasets comprising five years of hourly observations: two collected from independent system operators (ISOs) and one from a utility provider. Each dataset includes load demand measurements and covariate information. The ISO New England (ISO-NE) dataset spans 2015--2019 and contains day-ahead load demand estimates provided by the system operator, along with dry bulb temperature and dew point readings. The ``North American Utility'' (NAU) dataset is from an anonymous North American utility provider, and includes temperature readings from 1987 to 1991. The California ISO (CAISO) dataset contains information collected from 2021 to 2025 about fuel mixes from solar and wind generation, as well as average temperature readings across five regions in California to ensure statewide coverage. Load and fuel mix data were obtained from \texttt{\href{https://gridstatus.io/}{gridstatus.io}}, and temperature readings were sourced from the ERA5 dataset~\cite{era}. Table~\ref{tab:dataset_covs} summarizes the list of covariates available for each dataset.

\subsection{Data Splits}

For model training and evaluation, each dataset is partitioned into approximately 3.5 years of training data, 6 months of validation data, and 1 year of testing data, each at an hourly resolution. Formally, let $T$ denote the total number of observations, and let $N_{\text{yr}}$ and $N_{\text{6mo}}$ denote the number of hourly observations corresponding to one year and six months, respectively. The validation and test boundaries are defined as
\begin{align*}
    T_{\text{test}} = T - N_{\text{yr}} - L, \qquad
    T_{\text{val}} = T_{\text{test}} - N_{\text{6mo}},
\end{align*}
which yields the splits
\[
\begin{gathered}
\mathcal{T}_{\text{train}} = \{1,\ldots,T_{\text{val}}\}, \qquad
\mathcal{T}_{\text{val}} = \{T_{\text{val}}+1,\ldots,T_{\text{test}}\}, \\
\mathcal{T}_{\text{test}} = \{T_{\text{test}}+1,\ldots,T\}.
\end{gathered}
\]
Subtracting the context length in the test boundary ensures that exactly one year of load demand is forecast after accounting for the context required by each model.

\subsection{Noise Generation}
\label{subseq:noise}

Since the presence of future covariate values can naturally improve the quality of load forecasts, we consider the scenario in which a system operator knows a noisy forecast of covariate values for the following day-ahead forecast. For each dataset, let $x_{i,t}$ denote the value of the $i$th covariate at time $t$. We employ the following procedure for generating noise:
\begin{enumerate}
    \item Fit an ARIMA$(p,d,q)$ model on $\{x_{i,t}\}_{t\in\mathcal{T}_{\text{train}}}$.
    \item Compute the residuals over the validation set $\varepsilon_{i,t} := x_{i,t} - \hat{x}_{i,t}$.
    \item Compute the standard deviation $\sigma_i$ over the residuals.
\end{enumerate}
At predict time, we add noise drawn from a zero-mean Gaussian distribution with standard deviation $\lambda\sigma_{i}$ to each covariate, where $\lambda \geq 0$ is the standard deviation scaling factor that controls how noisy the distribution we draw from is. The only exception to this procedure is for the \texttt{DA\_DEMD} covariate from the ISO-NE dataset since it is a forecast of the load demand itself. For all other covariates, values of $p,d,q$ were chosen through grid search by selecting the model that yielded the lowest mean absolute error (MAE) and mean absolute percentage error (MAPE) over the validation set.

This perturbation model is intended as a controlled representation of covariate forecast uncertainty. In practice, weather and renewable generation forecasts may exhibit temporal dependence, bias, and non-Gaussianity. We use independent zero-mean Gaussian perturbations as an interpretable baseline, with $\lambda$ controlling the magnitude of forecast uncertainty. This allows us to examine how forecasting models behave as future covariate information becomes increasingly unreliable.

\begin{table}[t]
\centering
\caption{Covariates from each dataset used to forecast load.}
\label{tab:dataset_covs}

\begin{tabularx}{\linewidth}{l|X}
\toprule
\textbf{Dataset} & \textbf{Covariates} \\
\midrule
ISO-NE &
Day-ahead demand (\texttt{DA\_DEMD}), dry bulb temperature (\texttt{DryBulb}), dew point (\texttt{DewPnt}) \\
\midrule
NAU &
Temperature (\texttt{temperature}) \\
\midrule
CAISO &
Solar mix (\texttt{solar}), wind mix (\texttt{wind}), regions 1--5 temperature
(\texttt{r1\_temp\_c}, \ldots, \texttt{r5\_temp\_c}) \\
\bottomrule
\end{tabularx}
\end{table}

\subsection{Hyperparameters, Training, and Inference}

Structural hyperparameters for the TFS models were selected via random search on the ISO-NE dataset and are listed in Table~\ref{tab:model_hyperparameters}. All models use a context length of $L=168$, a batch size of $B=64$, and are trained for up to 250 epochs with an early stopping patience of 20 epochs. Since each TFS model supports multivariate forecasting, we optimize a multi-task objective consisting of the MSE for the target load forecast and the average MSE across the remaining covariates, weighted equally to encourage shared representations while maintaining load forecasting accuracy. To mitigate distribution shift between the training and testing data, we adopt the Series Stationarization module from \cite{liuNonstationaryTransformersExploring2023}, which normalizes each time series instance to zero mean and unit variance before restoring the predictions to the original scale. When future covariate values are available, they are normalized using statistics computed only over the context window to prevent information leakage.

The efficiency of inference using pre-trained models enables us to evaluate each TSFM across various context lengths. Since TSFMs are not trained specifically for load forecasting, longer context lengths may allow them to extract domain-relevant temporal patterns. We consider context lengths beginning at 256 and incrementing in powers of 2 up to each model's maximum context length (cf. Table~\ref{tab:tsfm_comparison}), and select the model that yields the lowest errors. Because the TSFMs we consider support different input structures, we choose to evaluate the performance of each model under its available capabilities rather than restricting our study to a common input specification. We compute point predictions by taking the mean of the predicted quantile distribution. Among the models considered, we only fine-tune Chronos-2 due to its strong zero-shot performance, ability to incorporate both historical and future covariates, and native library support for fine-tuning. We perform full fine-tuning with an initial learning rate of $\eta_0 = 10^{-6}$ and batch size of $B=64$ for 1000 steps.

To capture temporal dependencies in our data, we incorporate a binary holiday indicator and hour-of-day, day-of-week, and month-of-year information into each model. For DLinear and PatchTST, these features are represented using standard cyclical encodings (i.e., sine and cosine transformations). In contrast, Informer, TimesNet, and Chronos-2 take as input the discrete values of these calendar features and learn continuous representations using temporal embedding layers.

\subsection{Metrics}

We use two metrics to evaluate the performance of each model in our experiments: mean absolute error (MAE) and mean absolute percentage error (MAPE). These are defined over the test set as 
$\mae := \frac{1}{|\mathcal{T}_{\text{test}}|} \sum_{t \in \mathcal{T}_{\text{test}}} |y_t - \hat{y}_t|$ and $\mape := \frac{1}{|\mathcal{T}_{\text{test}}|} \sum_{t \in \mathcal{T}_{\text{test}}} |\frac{y_t - \hat{y}_t}{y_t}|\times 100\%$.

\subsection{Partial Leakage}

One concern when benchmarking TSFMs is the possibility of \emph{partial leakage}, whereby a model has been pre-trained on the same data used for evaluation. Fortunately, the TSFMs considered in this work publicly disclose their pre-training corpora. Based on these disclosures, to the best of our knowledge, none of the datasets used in our experiments were included in the pre-training data.

\begin{table}[t]
\caption{Hyperparameters of the forecasting models.}
\label{tab:model_hyperparameters}
\centering
\small
\begin{tabular}{lll}
\toprule
Model & Hyperparameter & Value \\
\midrule
\multirow{2}{*}{DLinear}
 & Moving average kernel & 25 \\
 & Learning rate & $10^{-4}$ \\
\midrule
\multirow{7}{*}{PatchTST}
 & Model dimension & 128 \\
 & Hidden dimension & 512 \\
 & Encoder layers & 3 \\
 & Attention heads & 8 \\
 & Patch length & 12 \\
 & Stride & 12 \\
 & Learning rate & $10^{-5}$ \\
\midrule
\multirow{7}{*}{Informer}
 & Model dimension & 64 \\
 & Hidden dimension & 256 \\
 & Label length & 48 \\
 & Encoder layers & 1 \\
 & Decoder layers & 2 \\
 & Attention heads & 8 \\
 & Learning rate & $10^{-4}$ \\
\midrule
\multirow{6}{*}{TimesNet}
 & Model dimension & 64 \\
 & Hidden dimension & 256 \\
 & Encoder layers & 3 \\
 & Top $k$ amplitudes & 5 \\
 & Number of kernels & 3 \\
 & Learning rate & $10^{-5}$ \\
\bottomrule
\end{tabular}
\end{table}

\begin{table*}[!htb]
\centering
\caption{Results for the historical-information-only setting. Lowest errors are shown in \textbf{bold} and second-lowest errors are \underline{underlined}.}
\label{tab:past_only_results}
\resizebox{\linewidth}{!}{%
\begin{tabular}{llcccccccc}
\toprule
&
&
\multicolumn{3}{c}{\textbf{Univariate}}
&
\multicolumn{4}{c}{\textbf{Multivariate}}
&
\textbf{Covariate-Informed} \\
\cmidrule(lr){3-5}
\cmidrule(lr){6-9}
\cmidrule(lr){10-10}
\textbf{Dataset}
& \textbf{Metric}
& TimesFM 2.5 (ZS)
& Moirai 2.0 (ZS)
& TiRex (ZS)
& DLinear
& PatchTST
& Informer
& TimesNet
& Chronos-2 (ZS/FT) \\
\midrule

\multirow{2}{*}{ISO-NE}
& MAE
& 561.43 & 503.93 & 575.45
& 616.18 & 590.10 & 615.87 & 566.78
& \underline{496.55}/\textbf{464.40} \\
& MAPE
& 3.91 & 3.52 & 4.03
& 4.34 & 4.20 & 4.36 & 4.02
& \underline{3.44}/\textbf{3.23} \\
\midrule

\multirow{2}{*}{NAU}
& MAE
& 78.89 & 71.80 & 87.16
& 95.80 & 94.80 & 83.13 & 72.54
& \underline{66.08}/\textbf{63.15} \\
& MAPE
& 3.38 & 3.06 & 3.73
& 4.18 & 4.21 & 3.64 & 3.14
& \underline{2.86}/\textbf{2.69} \\
\midrule

\multirow{2}{*}{CAISO}
& MAE
& 732.25 & 646.17 & 730.06
& 842.78 & 769.56 & 897.38 & 698.20
& \underline{604.26}/\textbf{567.35} \\
& MAPE
& 3.27 & 2.85 & 3.27
& 3.76 & 3.44 & 3.87 & 3.08
& \underline{2.72}/\textbf{2.55} \\
\bottomrule
\end{tabular}}
\end{table*}

\section{Experiment Results}
\label{sec:exp_results}

In each experiment, we assume that historical load demand and past covariate values are known. For future covariate values, we consider two scenarios: (i) they are known exactly, or (ii) they are provided as noisy forecasts.

\subsection{Without Future Covariates}

Table~\ref{tab:past_only_results} summarizes the forecasting performance when only historical information is available and future covariate values are unavailable. Among the univariate TSFMs, which rely solely on historical load demand for forecasting, Moirai 2.0 consistently achieves the best performance across all datasets despite having substantially fewer parameters than TimesFM 2.5 and TiRex and a shorter maximum context length than TimesFM 2.5. These results suggest that neither larger model size nor longer context windows necessarily lead to superior performance on STLF tasks.

Among the multivariate TFS models, TimesNet consistently delivers the strongest performance across all datasets. This highlights the effectiveness of its 2D backbone in jointly modeling temporal variations across multiple variables. Notably, TimesNet substantially outperforms the Transformer-based architectures PatchTST and Informer, reinforcing previous findings that Transformer architectures are not universally optimal for time series forecasting.

Chronos-2, however, outperforms all competing models in both the zero-shot and fine-tuned settings. While the strong performance of the other TSFMs demonstrates the benefits of large-scale pre-training for downstream forecasting tasks, Chronos-2 further shows that effectively incorporating covariate information enables TSFMs to achieve state-of-the-art performance on STLF.

\begin{table}[ht]
\centering
\caption{Results for the future-covariate setting. Lowest errors are shown in \textbf{bold} and second-lowest errors are \underline{underlined}.}
\label{tab:future_results}
\resizebox{\linewidth}{!}{%
\begin{tabular}{lllccc}
\toprule
&
&
&
\multicolumn{2}{c}{\textbf{Multivariate}}
&
\textbf{Covariate-Informed} \\
\cmidrule(lr){4-5}
\cmidrule(lr){6-6}
\textbf{Dataset}
& \textbf{Metric}
& $\lambda$
& Informer
& TimesNet
& Chronos-2 (ZS/FT) \\
\midrule

\multirow{8}{*}{ISO-NE}
& \multirow{4}{*}{MAE}
& 0   & 317.92 & \underline{263.98} & 266.25/\textbf{229.82} \\
&
& 0.5 & 350.33 & \underline{272.57} & 303.12/\textbf{267.04} \\ 
&
& 1   & 431.35 & \textbf{300.64} & 379.80/\underline{337.32} \\
&
& 5   & 1230.29 & \textbf{707.22} & 1207.72/\underline{985.60} \\ 
\cmidrule(lr){2-6}
&
\multirow{4}{*}{MAPE}
& 0   & 2.37 & \underline{1.91} & 1.92/\textbf{1.67} \\
&
& 0.5 & 2.60 & \underline{1.98} & 2.21/\textbf{1.94} \\ 
&
& 1   & 3.19 & \textbf{2.19} & 2.80/\underline{2.46} \\
&
& 5   & 9.13 & \textbf{5.17} & 9.14/\underline{7.29} \\ 
\midrule

\multirow{8}{*}{NAU}
& \multirow{4}{*}{MAE}
& 0   & 61.53 & \underline{51.29} & 52.80/\textbf{45.32} \\
&
& 0.5 & 65.42 & \underline{53.13} & 55.59/\textbf{50.42} \\ 
&
& 1   & 74.08 & \textbf{56.79} & 59.67/\underline{59.25} \\
&
& 5   & 136.51 & \textbf{122.49} & \underline{125.05}/130.50 \\ 
\cmidrule(lr){2-6}
&
\multirow{4}{*}{MAPE}
& 0   & 2.72 & \underline{2.24} & 2.31/\textbf{1.99} \\
&
& 0.5 & 2.88 & \underline{2.32} & 2.43/\textbf{2.21} \\ 
&
& 1   & 3.23 & \textbf{2.47} & 2.62/\underline{2.58} \\
&
& 5   & 5.84 & \textbf{5.33} & \underline{5.47}/5.62 \\ 
\midrule

\multirow{8}{*}{CAISO}
& \multirow{4}{*}{MAE}
& 0   & 669.42 & \underline{549.18} & 550.00/\textbf{478.77} \\
&
& 0.5 & 704.01 & \underline{554.88} & 618.49/\textbf{522.97} \\ 
&
& 1   & 791.66 & \textbf{571.97} & 732.78/\underline{589.63} \\
&
& 5   & 1703.48 & \textbf{845.82} & 1266.90/\underline{865.81} \\ 
\cmidrule(lr){2-6}
&
\multirow{4}{*}{MAPE}
& 0   & 2.96 & \underline{2.40} & 2.47/\textbf{2.13} \\
&
& 0.5 & 3.11 & \underline{2.43} & 2.76/\textbf{2.32} \\ 
&
& 1   & 3.49 & \textbf{2.50} & 3.25/\underline{2.60} \\
&
& 5   & 7.42 & \textbf{3.72} & 5.65/\underline{3.75} \\ 
\bottomrule
\end{tabular}}
\end{table}

\subsection{With Future Covariates}

As discussed previously, TimesFM 2.5, Moirai 2.0, and TiRex are designed for univariate forecasting and therefore do not incorporate covariate information. Likewise, DLinear and PatchTST employ channel-independent architectures in which each variate is modeled independently using a shared set of model parameters, without explicitly capturing inter-variate dependencies. Consequently, these models cannot leverage future covariate information and are excluded from this comparison. Accordingly, only Informer, TimesNet, and Chronos-2 are evaluated in this setting.

\begin{figure*}[!t]
\centering
\includegraphics[width=0.967\textwidth]{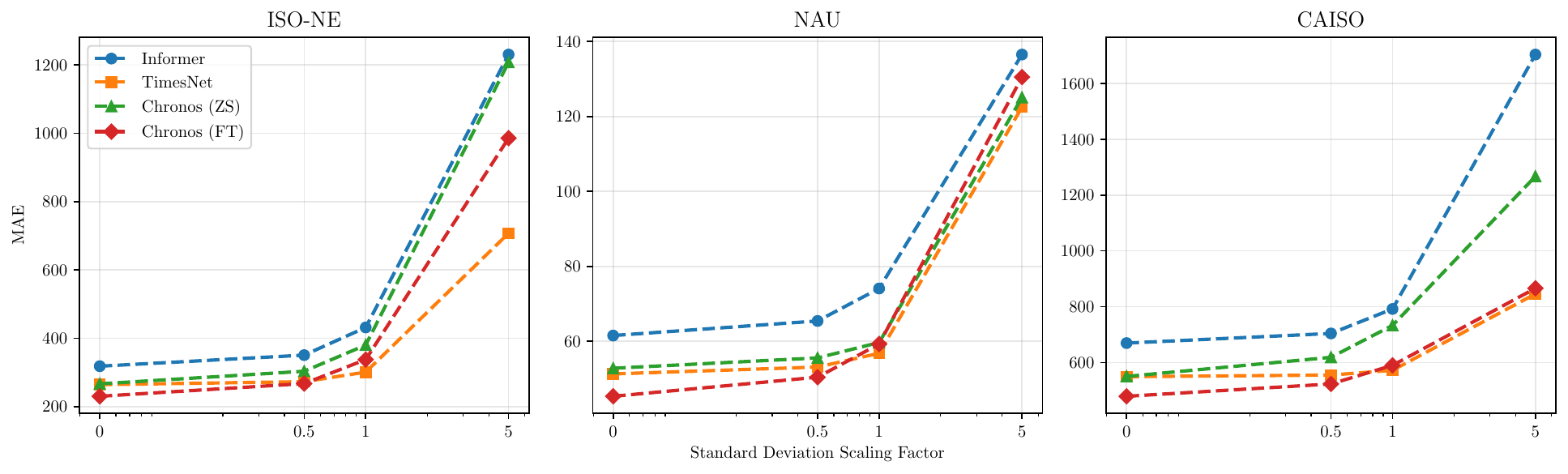}
\caption{Degradation comparison of each model across each dataset as variance of noise distribution increases.}
\label{fig:noisy_comparison}
\end{figure*}

Table~\ref{tab:future_results} summarizes the forecasting performance when historical observations are available and future covariate values are provided as forecasts with varying levels of uncertainty. The parameter $\lambda$ denotes the standard deviation scaling factor introduced in Section~\ref{subseq:noise}, where $\lambda=0$ corresponds to perfect knowledge of future covariates and $\lambda>0$ represents increasingly noisy covariate forecasts. When future covariates are noise-free ($\lambda=0$) or only mildly corrupted ($\lambda=0.5$), Chronos-2 achieves the best performance in the fine-tuned setting across all datasets, with TimesNet consistently ranking second. However, as the noise level increases, TimesNet surpasses Chronos-2 in both the zero-shot and fine-tuned settings, indicating greater robustness to uncertainty in future covariates. In contrast, Informer consistently exhibits the poorest performance across all datasets, with its accuracy deteriorating substantially as the covariate forecasts become noisier. Figure~\ref{fig:noisy_comparison} further illustrates the performance degradation of each model as the noise level increases.

TimesNet models temporal dependencies by identifying the top $k$ dominant periodicities in the input sequence and learning their corresponding temporal variations. The observed robustness of TimesNet under noisy covariate forecasts may be related to this frequency-based representation, as dominant periodic components could be less sensitive to random perturbations than the raw time-domain observations. This may allow TimesNet to retain useful temporal structure as covariate uncertainty increases. While this provides a plausible explanation for the observed behavior, the contribution of the frequency-based representation is not directly isolated in our experiments, and we relegate further analysis of this phenomenon to future work.

\section{Conclusion}
\label{sec:con}

In this paper, we presented a comprehensive benchmark of TFS models and TSFMs for short-term load forecasting under realistic operational scenarios with varying levels of future covariate availability and uncertainty. Across three real-world datasets, Chronos-2 consistently achieved state-of-the-art performance in both the zero-shot and fine-tuned settings, demonstrating the effectiveness of covariate-informed TSFMs for STLF. However, TimesNet consistently outperformed Chronos-2 when future covariate forecasts were highly noisy, suggesting that its frequency-based architecture is inherently more robust to covariate uncertainty. These findings highlight both the promise of foundation models for power system forecasting and the importance of robust covariate modeling in practical deployments. 

\bibliographystyle{ieeetr} 
\bibliography{references}

\end{document}